%% file: main.tex
\documentclass[11pt]{article}
\usepackage{acl2015}

\usepackage{latexsym}
\usepackage{url}
\usepackage{amsmath,amssymb}
\usepackage{ifthen}
\usepackage{graphicx}
\usepackage{stmaryrd}
\usepackage{enumitem}
\usepackage{multicol}
\usepackage{multirow}
\usepackage[usenames,dvipsnames,svgnames,table]{xcolor}
\usepackage{subfigure}
\usepackage{algorithm}
\usepackage{algorithmic}
\usepackage{xspace}
\usepackage[utf8]{inputenc}

\input std-macros

\input macros

\title{Compositional Semantic Parsing on Semi-Structured Tables}

\author{
  Panupong Pasupat \\
  Computer Science Department \\
  Stanford University \\
  {\tt ppasupat@cs.stanford.edu} \\\And
  Percy Liang \\
  Computer Science Department \\
  Stanford University \\
  {\tt pliang@cs.stanford.edu} \\}

\date{}

\begin{document}

\maketitle

\input{abstract}

\input{percyfigure}
\input{intro}

\input{task}

\input{approach}

\input{experiments}
\input{discussion}

\input{acknowledgement}
\bibliographystyle{acl}
\bibliography{all}

\end{document}

%% file: std-macros.tex
\newcommand\sD{\ensuremath{\mathcal{D}}}

\newcommand\sZ{\ensuremath{\mathcal{Z}}}

\newcommand\bR{\ensuremath{\mathbf{R}}}

\newcommand\FigTop[4]{\begin{figure}[t] \begin{center} \includegraphics[scale=#2]{#1} \end{center} \caption{\label{fig:#3} #4} \end{figure}}

\newcommand\interpret[1]{\llbracket #1 \rrbracket}

\newcommand\refeqn[1]{(\ref{eqn:#1})}

\newcommand\refsec[1]{Section~\ref{sec:#1}}

\newcommand\reffig[1]{Figure~\ref{fig:#1}}

\newcommand\reftab[1]{Table~\ref{tab:#1}}

\ifthenelse{\isundefined{\definition}}{\newtheorem{definition}{Definition}}{}
\ifthenelse{\isundefined{\assumption}}{\newtheorem{assumption}{Assumption}}{}
\ifthenelse{\isundefined{\proposition}}{\newtheorem{proposition}{Proposition}}{}
\ifthenelse{\isundefined{\theorem}}{\newtheorem{theorem}{Theorem}}{}
\ifthenelse{\isundefined{\lemma}}{\newtheorem{lemma}{Lemma}}{}
\ifthenelse{\isundefined{\corollary}}{\newtheorem{corollary}{Corollary}}{}
\ifthenelse{\isundefined{\alg}}{\newtheorem{alg}{Algorithm}}{}
\ifthenelse{\isundefined{\example}}{\newtheorem{example}{Example}}{}

%% file: macros.tex
\definecolor{gray}{gray}{0.5}

\renewcommand\FigTop[4]{\begin{figure}[t]\centering\includegraphics[scale=#2]{#1}\vspace*{-.6em}\caption{\label{fig:#3} #4} \end{figure}}

\newcommand\dataset{{\sc WikiTableQuestions}\xspace}
\newcommand\candidates[1][x]{\sZ_{#1}}
\newcommand\norm[1]{\left\| #1 \right\|}
\newcommand\T[1]{\text{\texttt{#1}}}
\newcommand\C[1]{\text{\emph{#1}}}
\newcommand\SC[1]{\text{\textsc{#1}}}
\newcommand\G[1]{\textcolor{gray}{(#1)}}

\newcommand\nl[1]{``\emph{#1}''}

%% file: abstract.tex
\begin{abstract}
Two important aspects of semantic parsing for question answering
are the breadth of the knowledge source and the depth of logical compositionality.
While existing work trades off one aspect for another,
this paper simultaneously makes progress on both fronts
through a new task:
answering complex questions on semi-structured tables
using question-answer pairs as supervision.
The central challenge
arises from two compounding factors:
the broader domain 
results in an open-ended set of relations,
and the deeper compositionality
results in a combinatorial explosion
in the space of logical forms.
We propose a logical-form driven parsing algorithm
guided by strong typing constraints
and show that it obtains
significant improvements over natural baselines.
For evaluation, we created a new dataset
of 22,033 complex questions on Wikipedia tables,
which is made publicly available.
\end{abstract}

%% file: percyfigure.tex
\begin{figure}[tb]\centering\small
\textsf{
\begin{tabular}{|l|l|l|l|} \hline
\textbf{Year} & \textbf{City} & \textbf{Country} & \textbf{Nations} \\ \hline
1896 & Athens & Greece & 14 \\
1900 & Paris & France & 24 \\
1904 & St.\ Louis & USA & 12 \\
\dots & \dots & \dots & \dots \\
2004 & Athens & Greece & 201 \\
2008 & Beijing & China & 204 \\
2012 & London & UK & 204 \\ \hline
\end{tabular}
}
\newline\vspace*{.1em}\newline
\newcommand{\myspace}{\rule{0pt}{1.2em}}
\begin{tabular}{@{\;}r@{: }p{7.0cm}} 
\myspace
$x_1$ & \emph{``Greece held its last Summer Olympics in which year?''} \\
$y_1$ & \{2004\} \\
\myspace
$x_2$ & \emph{``In which city's the first time with at least 20 nations?''} \\
$y_2$ & \{Paris\} \\
\myspace
$x_3$ & \emph{``Which years have the most participating countries?''} \\
$y_3$ & \{2008, 2012\} \\
\myspace
$x_4$ & \emph{``How many events were in Athens, Greece?''} \\
$y_4$ & \{2\} \\
\myspace
$x_5$ & \emph{``How many more participants were there in 1900 than in the first year?''} \\
$y_5$ & \{10\} \\
\end{tabular}
\caption{Our task is to answer a highly compositional question from an HTML table.
We learn a semantic parser from question-table-answer triples $\{(x_i,t_i,y_i)\}$.\vspace*{-.4em}}
\label{fig:example}
\end{figure}

%% file: intro.tex
\section{Introduction}
In semantic parsing for question answering,
natural language
questions
are converted
into logical forms,
which can
be executed on a knowledge source
to obtain answer denotations.
Early semantic parsing systems were trained to answer highly compositional questions,
but the knowledge sources were limited to small closed-domain databases
\cite{zelle96geoquery,wong07synchronous,zettlemoyer07relaxed,kwiatkowski11lex}.
More recent work sacrifices compositionality
in favor of
using more open-ended knowledge bases such as Freebase
\cite{cai2013large,berant2013freebase,fader2014open,reddy2014large}.
However,
even these broader knowledge sources still
define a rigid schema over entities and relation types,
thus restricting the scope of answerable questions.

To simultaneously increase both
the \emph{breadth} of the knowledge source
and the \emph{depth} of logical compositionality,
we propose a new task (with an associated dataset):
answering a question using an HTML table as the knowledge source.
\reffig{example} shows several question-answer pairs and an accompanying table,
which are typical of those in our dataset.
Note that the questions are logically quite complex,
involving a variety of operations such as
comparison ($x_2$), superlatives ($x_3$),
aggregation ($x_4$), and arithmetic ($x_5$).

The HTML tables are semi-structured and not normalized.
For example, a cell might contain multiple parts (e.g., \nl{Beijing, China} or \nl{200 km}).
Additionally, we mandate that the training and test tables are disjoint,
so at test time, we will see relations (column headers; e.g., \nl{Nations})
and entities (table cells; e.g., \nl{St.\ Louis})
that were not observed during training.
This is in contrast to knowledge bases like Freebase,
which have a global fixed relation schema with normalized entities and relations.

Our task setting produces two main challenges.
Firstly, the increased breadth in the knowledge source
requires us to generate logical forms from
novel tables with previously unseen relations and entities.
We therefore cannot
follow the typical semantic parsing strategy of
constructing or learning a lexicon that maps
phrases to relations ahead of time.
Secondly, the increased depth in compositionality
and additional logical operations
exacerbate the exponential growth of the number of possible logical forms.

We trained a semantic parser for this task from question-answer pairs
based on the framework illustrated in \reffig{framework}.
First, relations and entities from the semi-structured HTML table
are encoded in a graph.
Then, the system parses the question into candidate logical forms
with a high-coverage grammar,
reranks the candidates with a log-linear model,
and then executes the highest-scoring logical form
to produce the answer denotation.
We use beam search with pruning strategies
based on type and denotation constraints to control the
combinatorial explosion.

To evaluate the system, we created a new dataset, \dataset,
consisting of 2,108 HTML tables from Wikipedia and 22,033 question-answer pairs.
When tested on unseen tables, the system achieves an accuracy of 37.1\%,
which is significantly higher than the information retrieval baseline of 12.7\%
and a simple semantic parsing baseline of 24.3\%.

%% file: task.tex
\section{Task}

Our task is as follows:
given a table $t$
and a question $x$ about the table,
output a list of values $y$
that answers the question according to the table.
Example inputs and outputs are shown in \reffig{example}.
The system has access to
a training set $\sD = \{(x_i, t_i, y_i)\}_{i=1}^N$
of questions, tables, and answers,
but the tables in test data do not appear during training.

The only restriction on the question $x$
is that a person must be able to answer it
using just the table $t$.
Other than that, the question can be of any type,
ranging from a simple table lookup question
to a more complicated one that involves
various logical operations.

\input dataset

%% file: dataset.tex
\paragraph{Dataset.}
We created a new dataset, \dataset,
of question-answer pairs on HTML tables as follows.
We randomly selected data tables from Wikipedia with at least
8 rows and 5 columns.
We then created two Amazon Mechanical Turk tasks.
The first task asks workers to write trivia questions about the table.
For each question, we put one of the 36 generic prompts such as
\emph{``The question should require calculation''}
or \emph{``contains the word \emph{`first'} or its synonym''}
to encourage more complex utterances.
Next, we submit the resulting questions to the second task
where the workers answer each question based on the given table.
We only keep the answers that are agreed upon by at least two workers.
After this filtering, approximately 69\% of the questions remains.

The final dataset contains 22,033 examples on 2,108 tables.
We set aside 20\% of the tables 
and their associated questions
as the test set
and develop on the remaining examples.
Simple preprocessing was done on the tables:
We omit all non-textual contents of the tables,
and if there is a merged cell spanning many rows or columns,
we unmerge it and duplicate its content into each unmerged cell.
\refsec{x-dataset} analyzes various aspects of the dataset
and compares it to other datasets.

%% file: approach.tex
\section{Approach}
\FigTop{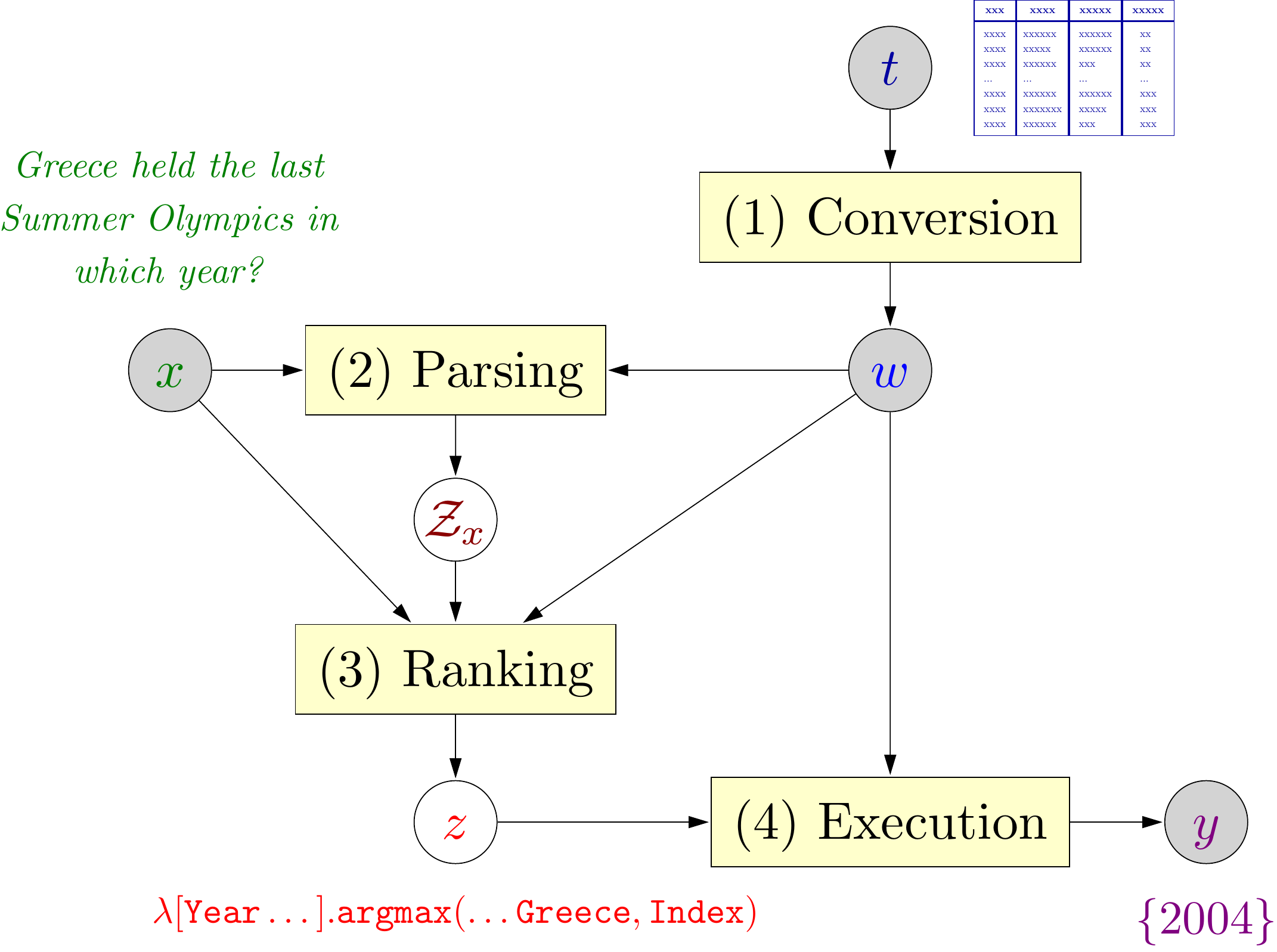}{0.32}{framework}
{The prediction framework:
(1) the table $t$ is deterministically converted into a knowledge graph $w$
as shown in \reffig{knowledgeGraph};
(2) with information from $w$,
the question $x$ is parsed into candidate logical forms in $\candidates$;
(3) the highest-scoring candidate $z\in\candidates$ is chosen; and
(4) $z$ is executed on $w$, yielding the answer $y$.
}

We now describe our semantic parsing framework
for answering a given question and for training the model
with question-answer pairs.

\textbf{Prediction.} Given a table $t$ and a question $x$,
we predict an answer $y$ using the framework illustrated in \reffig{framework}.
We first convert the table $t$ into a \emph{knowledge graph} $w$ (``world'')
which encodes different relations in the table (\refsec{graph}).
Next, we generate a set of candidate logical forms $\candidates$
by parsing the question $x$ using the information from $w$
(\refsec{parsingalgo}).
Each generated logical form $z \in \candidates$
is a graph query
that can be executed
on the knowledge graph $w$
to get a \emph{denotation} $\interpret{z}_w$.
We extract a feature vector $\phi(x,w,z)$ for each $z\in\candidates$ (\refsec{features})
and define a log-linear distribution over the candidates:
\begin{align}
p_\theta(z\mid x,w) \propto \exp\{\theta^\top \phi(x,w,z)\},
\end{align}
where $\theta$ is the parameter vector.
Finally, we choose the logical form $z$ with the highest model probability
and execute it on $w$ to get the answer denotation $y = \interpret{z}_w$.

\textbf{Training.}
Given training examples $\sD = \{(x_i, t_i, y_i)\}_{i=1}^N$,
we seek a parameter vector $\theta$ that maximizes the regularized log-likelihood of
the correct denotation $y_i$ marginalized over logical forms $z$.
Formally, we maximize the objective function
\begin{align}
J(\theta) = \frac{1}{N}\sum_{i=1}^N \log p_\theta(y_i\mid x_i, w_i) - \lambda\norm{\theta}_1,
\end{align}
where $w_i$ is deterministically generated from $t_i$, and
\begin{align}
p_\theta(y\mid x, w) = \sum_{z\in\candidates; y = \interpret{z}_w} p_\theta(z\mid x, w).
\end{align}

We optimize $\theta$ using AdaGrad \cite{duchi10adagrad},
running 3 passes over the data.
We use $L_1$ regularization
with $\lambda = 3\times 10^{-5}$ obtained from cross-validation.

The following sections explain individual system components in more detail.

\input graph

\input lambdadcs

\input parsing

%% file: graph.tex
\section{Knowledge graph}\label{sec:graph}

\FigTop{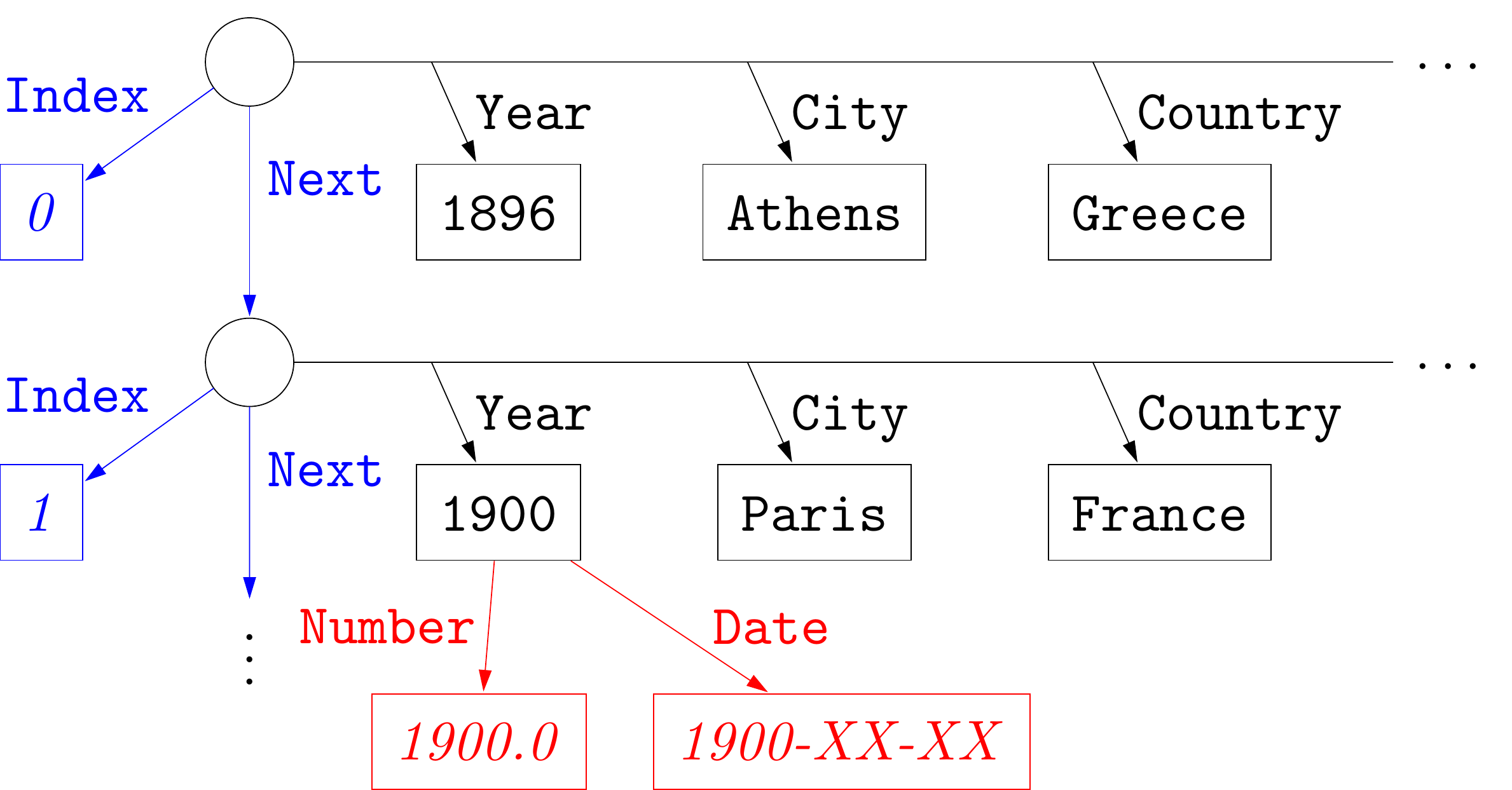}{0.30}{knowledgeGraph}
{Part of the knowledge graph corresponding to the table in \reffig{example}.
Circular nodes are row nodes.
We augment the graph with different entity normalization nodes
such as $\T{Number}$ and $\T{Date}$ (red)
and additional row node relations $\T{Next}$ and $\T{Index}$ (blue).}

Inspired by the graph representation of knowledge bases,
we preprocess the table $t$
by deterministically converting it into a \emph{knowledge graph} $w$
as illustrated in \reffig{knowledgeGraph}.
In the most basic form,
table rows become row nodes,
strings in table cells become entity nodes,\footnote{Two occurrences of the same string constitute one node.}
and table columns become directed edges
from the row nodes to the entity nodes of that column.
The column headers are used as edge labels for these row-entity relations.

The knowledge graph representation is convenient for three reasons.
First, we can
encode different forms of entity normalization in the graph.
Some entity strings (e.g., \emph{``1900''}) can be interpreted as a number, a date,
or a proper name depending on the context,
while some other strings
(e.g., \emph{``200 km''})
have multiple parts.
Instead of committing to one normalization scheme,
we introduce edges
corresponding to different normalization methods from the entity nodes.
For example, the node $\T{1900}$ will have an edge called $\T{Date}$
to another node \emph{1900-XX-XX} of type date.
Apart from type checking,
these normalization nodes also aid learning by
providing signals on the appropriate answer type.
For instance,
we can define a feature that associates the phrase \emph{``how many''}
with a logical form that says
``traverse a row-entity edge, then a $\T{Number}$ edge''
instead of just ``traverse a row-entity edge.''

The second benefit of the graph representation is its ability to
handle various logical phenomena via graph augmentation.
For example, to answer questions of the form
\emph{``What is the next \dots ?''}
or \emph{``Who came before \dots ?''},
we augment each row node with an edge labeled $\T{Next}$
pointing to the next row node,
after which the questions can be answered by traversing the $\T{Next}$ edge.
In this work, we choose to add two special edges on each row node:
the $\T{Next}$ edge mentioned above
and an $\T{Index}$ edge pointing to the row index number ($\C{0}, \C{1}, \C{2}, \dots$).

Finally,
with a graph representation,
we can query it directly using a logical formalism for knowledge graphs,
which we turn to next.

%% file: lambdadcs.tex
\section{Logical forms}\label{sec:lambdadcs}
As our language for logical forms,
we use
lambda dependency-based compositional semantics
\cite{liang2013lambdadcs},
or lambda DCS,
which we briefly describe here.
Each lambda DCS logical form is either a \emph{unary}
(denoting a list of values)
or a \emph{binary}
(denoting a list of pairs).
The most basic unaries are singletons
(e.g., $\T{China}$ represents an entity node,
and $\C{30}$ represents a single number),
while the most basic binaries are relations
(e.g., $\T{City}$ maps rows to city entities,
$\T{Next}$ maps rows to rows,
and $\T{>=}$ maps numbers to numbers).
Logical forms can be combined
into larger ones
via various operations
listed in \reftab{lambdadcs}.
Each operation produces a unary
except lambda abstraction:
$\lambda x[f(x)]$ is a binary mapping $x$ to $f(x)$.

\begin{table}[tb]\centering\small
\begin{tabular}{@{\;}ll@{}}
\textbf{Name} & \textbf{Example} \\ \hline

Join & $\T{City}.\T{Athens}$ \\
& \G{row nodes with a \T{City} edge to \T{Athens}} \\
Union & $\T{City}.(\T{Athens} \sqcup \T{Beijing})$ \\
Intersection & $\T{City}.\T{Athens} \sqcap \T{Year}.\T{Number}.\T{<}.\C{1990}$ \\
Reverse & $\bR[\T{Year}].\T{City}.\T{Athens}$ \\
\multicolumn{2}{l}{\G{entities where a row in $\T{City}.\T{Athens}$ has a \T{Year} edge to}} \\
Aggregation & $\T{count}(\T{City}.\T{Athens})$ \\
& \G{the number of rows with city \T{Athens}} \\
Superlative & $\T{argmax}(\T{City}.\T{Athens}, \T{Index})$ \\
& \G{the last row with city \T{Athens}} \\
Arithmetic & $\T{sub}(\C{204}, \C{201})$ \quad \G{$=204 - 201$} \\
Lambda & $\lambda x[\T{Year}.\T{Date}.x]$ \\
& \G{a binary: composition of two relations}\\

\hline

\end{tabular}
\caption{
The lambda DCS operations we use.
}\label{tab:lambdadcs}
\end{table}

%% file: parsing.tex
\section{Parsing and ranking}\label{sec:parsing}
\input rulescrumbs

Given the knowledge graph $w$,
we now describe how to parse the utterance $x$ into a set of candidate
logical forms $\candidates$

\subsection{Parsing algorithm}\label{sec:parsingalgo}

We propose a new \emph{floating parser} which is more flexible than a standard
chart parser.  Both parsers recursively build up derivations
and corresponding logical forms by repeatedly applying deduction rules,
but the floating parser allows logical form predicates
to be generated independently from the utterance.

\textbf{Chart parser.}
We briefly review the CKY algorithm for chart parsing to introduce notation.
Given an utterance with tokens $x_1,\dots,x_n$,
the CKY algorithm applies deduction rules of the following two kinds:
\begin{align}
  &(\C{TokenSpan}, i, j)[s] \to (c, i, j)[f(s)], \label{eqn:ckyBase} \\
  &(c_1,i,k)[z_1] + (c_2,k+1,j)[z_2] & \\
  &\hspace{1.17in} \to (c,i,j)[f(z_1,z_2)]. \nonumber
\end{align}
The first rule is a lexical rule
that matches an utterance token span $x_i \cdots x_j$ (e.g., $s = \emph{``New York''}$)
and produces a logical form (e.g., $f(s) = \T{NewYorkCity}$) with category $c$ (e.g., $\T{Entity}$).
The second rule takes two adjacent spans giving rise to 
logical forms $z_1$ and $z_2$ and builds a new logical form $f(z_1,z_2)$.
Algorithmically,
CKY stores derivations of category $c$ covering the span $x_i \cdots x_j$
in a \emph{cell} $(c,i,j)$.
CKY fills in the cells 
of increasing span lengths, and
the logical forms in the top cell $(\C{ROOT},1,n)$ are returned.

\firstRulesCrumb
\secondRuleCrumb

\textbf{Floating parser.}
Chart parsing uses lexical rules \refeqn{ckyBase}
to generate relevant logical predicates,
but in our setting of semantic parsing on tables,
we do not have the luxury of starting with or inducing a full-fledged lexicon.
Moreover, there is a mismatch between words in the utterance and
predicates in the logical form.
For instance, consider the question
\emph{``Greece held its last Summer Olympics in which year?''}
on the table in \reffig{example}
and the correct logical form
{\small $\bR[\lambda x[\T{Year}.\T{Date}.x]].\T{argmax}(\T{Country}.\T{Greece},\T{Index})$}.
While the entity $\T{Greece}$ can be anchored to the token \emph{``Greece''},
some logical predicates (e.g., $\T{Country}$)
cannot be clearly anchored to a token span.
We could potentially learn to
anchor the logical form $\T{Country}.\T{Greece}$ to \emph{``Greece''},
but if the relation $\T{Country}$ is not seen during training,
such a mapping is impossible to learn from the training data.
Similarly, some prominent tokens (e.g., \emph{``Olympics''})
are irrelevant and have no predicates anchored to them.

Therefore, instead of anchoring each predicate in the logical form
to tokens in the utterance via lexical rules,
we propose parsing more freely.
We replace the
anchored cells $(c,i,j)$ with \emph{floating cells} $(c,s)$
of category $c$ and logical form size $s$.
Then we apply rules of the following three kinds:
\begin{align}
  &(\C{TokenSpan}, i, j)[s] \to (c, 1)[f(s)], \label{eqn:anchoredBase} \\
  &\hspace{1.1in} \emptyset \to (c, 1)[f()], \label{eqn:floatingBase} \\
  &(c_1,s_1)[z_1] + (c_2,s_2)[z_2] & \label{eqn:floatingComp} \\
  &\hspace{1.17in} \to (c,s_1+s_2+1)[f(z_1,z_2)]. \nonumber
\end{align}
Note that rules \refeqn{anchoredBase} are similar to \refeqn{ckyBase} in chart parsing
except that the floating cell $(c,1)$
only keeps track of the category and its size $1$, not the span $(i,j)$.
Rules \refeqn{floatingBase} allow us to construct predicates out of thin air.
For example, we can construct a logical form representing a table relation $\T{Country}$
in cell $(\C{Relation}, 1)$ using the rule $\emptyset \to \C{Relation}\,[\T{Country}]$
independent of the utterance.
Rules \refeqn{floatingComp} perform composition,
where the induction is on the size $s$ of the logical form rather than the span length.
The algorithm stops when the specified maximum size is reached,
after which the logical forms in cells $(\C{ROOT}, s)$ for any $s$
are included in $\candidates$.
\reffig{parseTree} shows an example derivation generated by our floating parser.

The floating parser is very flexible: it can skip tokens
and combine logical forms in any order.
This flexibility might seem too unconstrained,
but we can use strong typing constraints to prevent nonsensical derivations from being constructed.

\FigTop{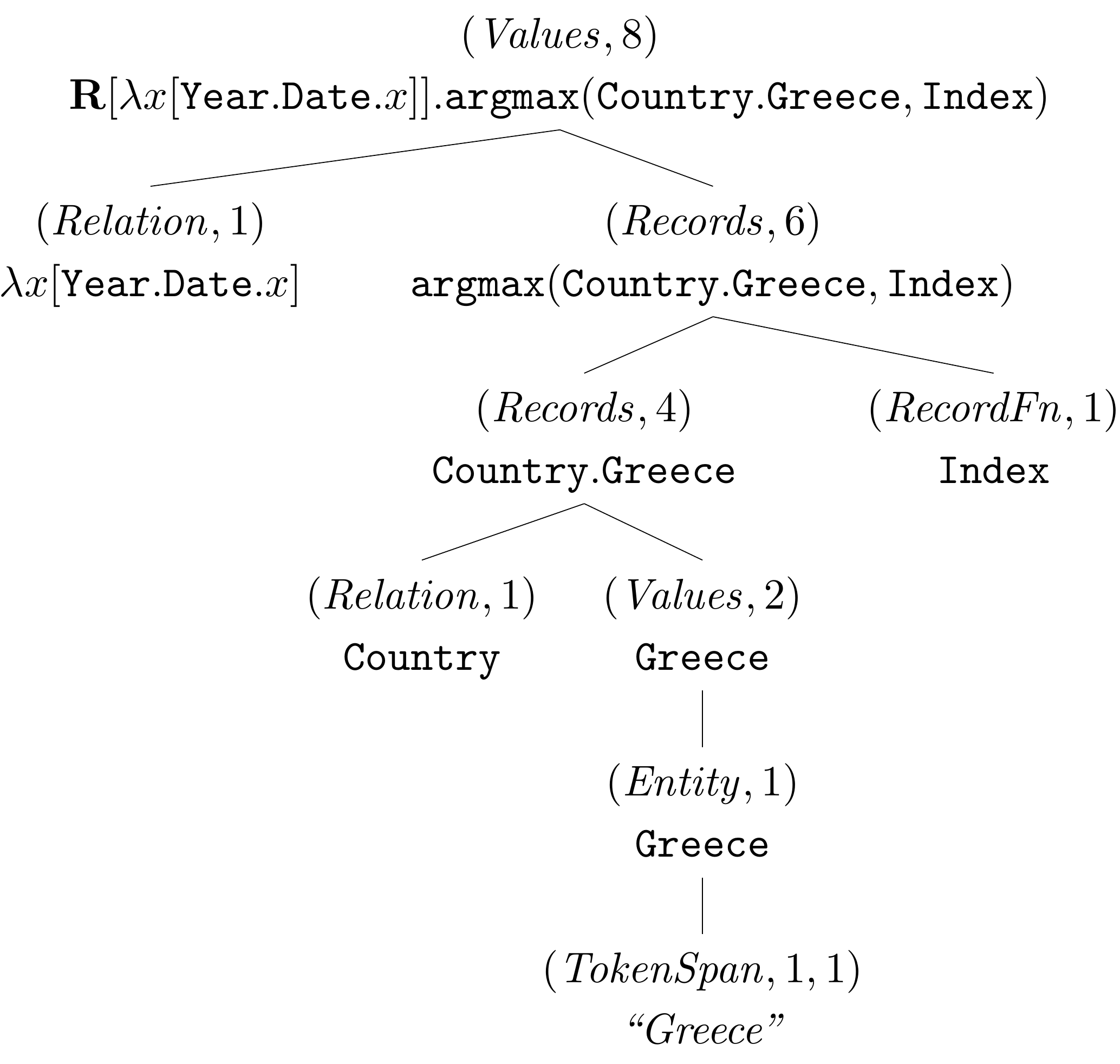}{0.35}{parseTree}
{A derivation for the utterance
\emph{``Greece held its last Summer Olympics in which year?''}
Only $\T{Greece}$ is anchored to a phrase \emph{``Greece''};
$\T{Year}$ and other predicates are floating.}

Tables~\ref{tab:rulesBase}~and~\ref{tab:rulesCompose}
show the full set of deduction rules we use.
We assume that all named entities
will explicitly appear in the question $x$,
so we anchor all entity predicates (e.g., $\T{Greece}$)
to token spans (e.g., \emph{``Greece''}).
We also anchor all numerical values (numbers, dates, percentages, etc.)
detected by an NER system.
In contrast, relations (e.g., $\T{Country}$)
and operations (e.g., $\T{argmax}$)
are kept floating since we want to learn how they are expressed in language.
Connections between phrases in $x$ and the generated 
relations and operations in $z$
are established in the ranking model through features.

\input features

\subsection{Generation and pruning}\label{sec:pruning}
Due to their recursive nature,
the rules allow us to generate highly compositional logical forms.
However, the compositionality comes at the cost of generating
exponentially many logical forms, most of which are redundant
(e.g., logical forms with an $\T{argmax}$ operation on a set of size 1).
We employ several methods to deal with this combinatorial explosion:

\textbf{Beam search.}
We compute the model probability of each partial logical form
based on available features (i.e., features that do not depend on the final denotation)
and keep only the $K=200$ highest-scoring logical forms in each cell.

\textbf{Pruning.}
We prune partial logical forms that lead
to invalid or redundant final logical forms.
For example, we eliminate any logical form that does not type check
(e.g., $\T{Beijing} \sqcup \T{Greece}$),
executes to an empty list
(e.g., $\T{Year}.\T{Number}.\C{24}$),
includes an aggregate or superlative on a singleton set
(e.g., $\T{argmax}(\T{Year}.\T{Number}.\C{2012}, \T{Index})$),
or joins two relations that are the reverses of each other
(e.g., $\bR[\T{City}].\T{City}.\T{Beijing}$).

%% file: rulescrumbs.tex
\newcommand\explain[1]{\multicolumn{3}{c}{\G{#1}}}
\newcommand\noexplain{\multicolumn{3}{c}{}}

\newcommand\firstRulesCrumb{
\begin{table}[tb]\centering\small
\begin{tabular}{@{\;}r@{ $\to$ }lll@{}}
\multicolumn{2}{c}{\textbf{Rule}} & \textbf{Semantics}
& \textbf{Example} \\ \hline

  \multicolumn{4}{c}{\textbf{\emph{Anchored to the utterance}}} \\

$\C{TokenSpan}$ & $\C{Entity}$
& $\mathrm{match}(z_1)$
& $\T{Greece}$ \\
\explain{$\mathrm{match}(s)$ = entity with name $s$}
&\hspace{-1.0em} anchored to \emph{``Greece''} \\

$\C{TokenSpan}$ & $\C{Atomic}$
& $\mathrm{val}(z_1)$
& $\C{2012-07-XX}$ \\
\explain{$\mathrm{val}(s)$ = interpreted value} 
&\hspace{-2.0em} anchored to \emph{``July 2012''} \\

\hline

\multicolumn{4}{c}{\textbf{\emph{Unanchored (floating)}}} \\

\multicolumn{2}{l}{\quad$\emptyset$ $\to$ $\C{Relation}$}
& $r$
& $\T{Country}$ \\
\explain{$r$ = row-entity relation} \\

\multicolumn{2}{l}{\quad$\emptyset$ $\to$ $\C{Relation}$}
& $\lambda x[r.p.x]$
& $\lambda x[\T{Year}.\T{Date}.x]$ \\
\explain{$p$ = normalization relation} \\

\multicolumn{2}{l}{\quad$\emptyset$ $\to$ $\C{Records}$}
& $\T{Type}.\T{Row}$
& \G{list of all rows} \\

\multicolumn{2}{l}{\quad$\emptyset$ $\to$ $\C{RecordFn}$}
& $\T{Index}$
& \G{row $\gets$ row index} \\

\hline

\end{tabular}
\caption{
Base deduction rules.
Entities and atomic values (e.g., numbers, dates) are anchored to token spans,
while other predicates are kept floating.
($a \gets b$ represents a binary mapping $b$ to $a$.)
}\label{tab:rulesBase}
\end{table}
}

\newcommand\secondRuleCrumb{
\begin{table*}[tb]\centering\small
\begin{tabular}{@{\;}r@{ $\to$ }lll@{\;}}
\multicolumn{2}{c}{\textbf{Rule}} & \textbf{Semantics}
& \textbf{Example} \\ \hline

\multicolumn{4}{c}{\textbf{\emph{Join + Aggregate}}} \\ 

$\C{Entity}$ or $\C{Atomic}$ & $\C{Values}$
& $z_1$
& $\T{China}$ \\

$\C{Atomic}$ & $\C{Values}$
& $c.z_1$
& $\T{>=}.\C{30}$
\quad\G{at least 30} \\
\explain{$c \in \{\T{<}, \T{>}, \T{<=}, \T{>=}\}$} \\

$\C{Relation} + \C{Values}$ & $\C{Records}$
& $z_1.z_2$
& $\T{Country}.\T{China}$
\quad\G{events (rows) where the country is China} \\

$\C{Relation} + \C{Records}$ & $\C{Values}$
& $\bR[z_1].z_2$
& $\bR[\T{Year}].\T{Country}.\T{China}$
\quad\G{years of events in China}\\

$\C{Records}$ & $\C{Records}$
& $\T{Next}.z_1$
& $\T{Next}.\T{Country}.\T{China}$
\quad\G{\dots before China}\\

$\C{Records}$ & $\C{Records}$
& $\bR[\T{Next}].z_1$
& $\bR[\T{Next}].\T{Country}.\T{China}$
\quad\G{\dots after China}\\

$\C{Values}$ & $\C{Atomic}$
& $a(z_1)$
& $\T{count}(\T{Country}.\T{China})$
\quad\G{How often did China \dots}\\
\explain{$a \in \{\T{count}, \T{max}, \T{min}, \T{sum}, \T{avg}\}$} \\

$\C{Values}$ & $\C{ROOT}$
& $z_1$ \\

\hline

\multicolumn{4}{c}{\textbf{\emph{Superlative}}} \\ 

$\C{Relation}$ & $\C{RecordFn}$
& $z_1$
& $\lambda x[\T{Nations}.\T{Number}.x]$
\quad\G{row $\gets$ value in Nations column} \\

$\C{Records} + \C{RecordFn}$ & $\C{Records}$
& $s(z_1, z_2)$
& $\T{argmax}(\T{Type}.\T{Row}, \lambda x[\T{Nations}.\T{Number}.x])$ \\

\explain{$s \in \{\T{argmax}, \T{argmin}\}$}
& \hspace*{1em}\G{events with the most participating nations} \\

\multicolumn{3}{c}{}
& $\T{argmin}(\T{City}.\T{Athens}, \T{Index})$
\quad\G{first event in Athens} \\

$\C{Relation}$ & $\C{ValueFn}$
& $\bR[\lambda x[a(z_1.x)]]$
& $\bR[\lambda x[\T{count}(\T{City}.x)]]$ 
\quad\G{city $\gets$ num. of rows with that city} \\

$\C{Relation} + \C{Relation}$ & $\C{ValueFn}$
& $\lambda x[\bR[z_1].z_2.x]$
& $\lambda x[\bR[\T{City}].\T{Nations}.\T{Number}.x]$ \\
\multicolumn{3}{c}{}
&\quad\G{city $\gets$ value in Nations column} \\

$\C{Values} + \C{ValueFn}$ & $\C{Values}$
& $s(z_1, z_2)$
& $\T{argmax}(\dots, \bR[\lambda x[\T{count}(\T{City}.x)]])$
\quad\G{most frequent city} \\

\hline

\multicolumn{4}{c}{\textbf{\emph{Other operations}}} \\ 
{\scriptsize$\C{ValueFn} + \C{Values} + \C{Values}$}

& $\C{Values}$
& \hspace*{-1em}{\scriptsize $o(\bR[z_1].z_2,\bR[z_1].z_3)$}
& $\T{sub}(\bR[\T{Number}].\bR[\T{Nations}].\T{City}.\T{London}, \dots)$ \\

\explain{$o \in \{\T{add}, \T{sub}, \T{mul}, \T{div}\}$}
& \hspace*{1em}\G{How many more participants were in London than \dots} \\

$\C{Entity} + \C{Entity}$ & $\C{Values}$
& $z_1 \sqcup z_2$
& $\T{China} \sqcup \T{France}$
\quad\G{China or France} \\

$\C{Records} + \C{Records}$ & $\C{Records}$
& $z_1 \sqcap z_2$
& $\T{City}.\T{Beijing} \sqcap \T{Country}.\T{China}$
\quad\G{\dots in Beijing, China} \\

\hline

\end{tabular}
\caption{
Compositional deduction rules.
Each rule $c_1, \dots, c_k \to c$ takes logical forms $z_1, \dots, z_k$
constructed over categories $c_1, \dots, c_k$, respectively,
and produces a logical form based on the semantics.
}\label{tab:rulesCompose}
\end{table*}
}

%% file: features.tex
\subsection{Features}\label{sec:features}

\begin{table}[tb]\centering\small
\makebox[7cm]{
\emph{``Greece held its last Summer Olympics in which year?''}
} \\
$z = \bR[\lambda x[\T{Year}.\T{Number}.x]].\T{argmax}(\T{Type}.\T{Row},\T{Index})$ \\
$y = \{\text{\emph{2012}}\}$ (type: \textsc{Num}, column: \textsc{Year}) \\
\vspace{.5em}
\begin{tabular}{@{ }ll@{ }}
\textbf{Feature Name} & \textbf{Note} \\ \hline
\arrayrulecolor{gray!25}
$(\C{``last''},\text{predicate} = \T{argmax})$ & lex \\
$\text{phrase} = \text{predicate}$ & unlex ($\because \C{``year''} = \T{Year}$) \\
\hline
missing entity & unlex ($\because$ missing \emph{Greece}) \\
\hline
$\text{denotation type} = \SC{Num}$ \\
$\text{denotation column} = \SC{Year}$ \\
\hline
$(\C{``which year''},\text{type} = \SC{Num})$ & lex \\
$\text{phrase} = \text{column}$ & unlex ($\because \C{``year''} = \SC{Year}$) \\
\hline
$(Q = \C{``which''},\text{type} = \SC{Num})$ & lex \\
$(H = \C{``year''},\text{type} = \SC{Num})$ & lex \\
$H = \text{column}$ & unlex ($\because \C{``year''} = \SC{year}$) \\
\arrayrulecolor{black}
\hline
\end{tabular}
\caption{
Example features that fire for the (incorrect) logical form $z$.
All features are binary.
(lex = lexicalized)
}\label{tab:features}
\end{table}

We define features $\phi(x,w,z)$ for our log-linear model
to capture the relationship between the question $x$ and the candidate $z$.
\reftab{features} shows some example features from each feature type.
Most features are of the form $(f(x), g(z))$ or $(f(x), h(y))$
where $y = \interpret{z}_w$ is the denotation,
and $f$, $g$, and $h$ extract some information (e.g., identity, POS tags)
from $x$, $z$, or $y$, respectively.

\textbf{phrase-predicate:}
Conjunctions between n-grams $f(x)$ from $x$
and predicates $g(z)$ from $z$.
We use both lexicalized features,
where all possible pairs $(f(x), g(z))$ form distinct features,
and binary unlexicalized features indicating whether
$f(x)$ and $g(z)$ have a string match.

\textbf{missing-predicate:}
Indicators on whether there are entities or relations
mentioned in $x$ but not in $z$.
These features are unlexicalized.

\textbf{denotation:}
Size and type of the denotation $y = \interpret{x}_w$.
The type can be either a primitive type (e.g., $\SC{Num}$, $\SC{Date}$, $\SC{Entity}$)
or the name of the column containing the entity in $y$ (e.g., $\SC{City}$).

\textbf{phrase-denotation:}
Conjunctions between n-grams from $x$ and the types of $y$.
Similar to the phrase-predicate features, we use both lexicalized and unlexicalized features.

\textbf{headword-denotation:}
Conjunctions between the question word $Q$
(e.g., \emph{what}, \emph{who}, \emph{how many})
or the headword $H$ (the first noun after the question word) with the types of $y$.

%% file: experiments.tex
\section{Experiments}\label{sec:experiments}

\subsection{Main evaluation}\label{sec:x-main}
We evaluate the system on the development sets
(three random 80:20 splits of the training data)
and the test data.
In both settings, the tables we test on do not appear during training.

\textbf{Evaluation metrics.}
Our main metric is \emph{accuracy},
which is the number of examples $(x,t,y)$
on which the system outputs the correct answer $y$.
We also report the \emph{oracle} score,
which counts the number of examples
where at least one generated candidate
$z\in\candidates$ executes to $y$.

\textbf{Baselines.}
We compare the system to two baselines.
The first baseline (IR), which simulates information retrieval,
selects an answer $y$ among the entities in the table
using a log-linear model over entities (table cells) rather than logical forms.
The features are conjunctions between
phrases in $x$ and properties of the answers $y$,
which cover all features in our main system that do not involve the logical form.
As an upper bound of this baseline,
69.1\% of the development examples
have the answer appearing as an entity in the table.

In the second baseline (WQ), we only allow deduction rules
that produce join and count logical forms.
This rule subset has the same logical coverage as
\newcite{berant2014paraphrasing},
which is designed to handle the 
\textsc{WebQuestions} \cite{berant2013freebase}
and
\textsc{Free917} \cite{cai2013large} datasets.

\begin{table}[tb]\centering
\begin{tabular}{lrrrr}
& \multicolumn{2}{c}{\textbf{dev}}
& \multicolumn{2}{c}{\textbf{test}} \\[-.3em]
& \textbf{acc} & \textbf{ora} & \textbf{acc} & \textbf{ora} \\ \hline
IR baseline & 13.4 & 69.1 & 12.7 & 70.6 \\
WQ baseline & 23.6 & 34.4 & 24.3 & 35.6 \\
Our system & 37.0 & 76.7 & 37.1 & 76.6 \\ \hline
\end{tabular}
\caption{Accuracy (acc) and oracle scores (ora) on the development sets
(3 random splits of the training data)
and the test data.}\label{tab:x-main}
\end{table}
\begin{table}[tb]\centering\small
\begin{tabular}{@{}r@{ }lrr}
&& \textbf{acc} & \textbf{ora} \\ \hline
& \textbf{Our system} & 37.0 & 76.7 \\ \hline
(a) & \multicolumn{3}{l}{\textbf{Rule Ablation}} \\
& join only & 10.6 & 15.7 \\
& join + count (= WQ baseline) & 23.6 & 34.4 \\
& join + count + superlative & 30.7 & 68.6 \\
& all $-$ $\{\sqcap, \sqcup\}$ & 34.8 & 75.1 \\
\hline
(b) & \multicolumn{3}{l}{\textbf{Feature Ablation}} \\
& all $-$ features involving predicate & 11.8 & 74.5 \\
& \quad all $-$ phrase-predicate & 16.9 & 74.5 \\
& \qquad all $-$ lex phrase-predicate & 17.6 & 75.9 \\
& \qquad all $-$ unlex phrase-predicate & 34.3 & 76.7 \\
& \quad all $-$ missing-predicate & 35.9 & 76.7 \\
& all $-$ features involving denotation & 33.5 & 76.8 \\
& \quad all $-$ denotation & 34.3 & 76.6 \\
& \quad all $-$ phrase-denotation & 35.7 & 76.8 \\
& \quad all $-$ headword-denotation & 36.0 & 76.7 \\
\hline
(c) & \multicolumn{1}{l}{\textbf{Anchor operations to trigger words}} & 37.1 & 59.4 \\ \hline
\end{tabular}
\caption{Average accuracy and oracle scores
on development data in various system settings.}\label{tab:x-ablation}
\end{table}

\textbf{Results.}
\reftab{x-main} shows the results compared to the baselines.
Our system gets an accuracy of 37.1\% on the test data,
which is significantly higher than both baselines,
while the oracle is 76.6\%.
The next subsections analyze the system components in more detail.

\subsection{Dataset statistics}\label{sec:x-dataset}
In this section, we analyze the breadth and depth
of the \dataset dataset,
and how the system handles them.

\textbf{Number of relations.}
With 3,929 unique column headers (relations)
among 13,396 columns,
the tables in the \dataset dataset contain many more relations than 
closed-domain datasets such as Geoquery \cite{zelle96geoquery}
and ATIS \cite{price1990atis}.
Additionally,
the logical forms that execute to the correct denotations
refer to a total of 2,056 unique column headers,
which is greater than the number of relations in
the \textsc{Free917} dataset (635 Freebase relations).

\textbf{Knowledge coverage.}
We sampled 50 examples from the dataset
and tried to answer them manually using Freebase.
Even though Freebase contains some information extracted from Wikipedia,
we can answer only 20\% of the questions,
indicating that \dataset\ contains a broad set of facts beyond Freebase.

\textbf{Logical operation coverage.}
The dataset covers a wide range of question types and logical operations.
\reftab{x-ablation}(a) shows the drop in oracle scores when
different subsets of rules are used to generate candidates logical forms.
The \emph{join only} subset corresponds to simple table lookup,
while \emph{join + count} is the WQ baseline
for Freebase question answering
on the \textsc{WebQuestions} dataset.
Finally, \emph{join + count + superlative} roughly corresponds to
the coverage of the Geoquery dataset.

\begin{table}[tb]\centering\small
\begin{tabular}{lr}
\textbf{Operation} & \textbf{Amount} \\ \hline
join (table lookup) & 13.5\% \\
+ join with $\T{Next}$ & + 5.5\% \\
+ aggregate ($\T{count}$, $\T{sum}$, $\T{max}$, \dots) & + 15.0\% \\
+ superlative ($\T{argmax}$, $\T{argmin}$) & + 24.5\% \\
+ arithmetic, $\sqcap$, $\sqcup$ & + 20.5\% \\
+ other phenomena & + 21.0\% \\ \hline
\end{tabular}
\caption{The logical operations required to answer the questions
in 200 random examples.}\label{tab:x-question-types}
\end{table}

To better understand the distribution of logical operations
in the \dataset dataset,
we manually classified 200 examples based on the types of operations required
to answer the question.
The statistics in \reftab{x-question-types} shows that
while a few questions only require simple operations such as table lookup,
the majority of the questions demands more advanced operations.
Additionally, 21\% of the examples cannot be answered using any logical form
generated from the current deduction rules;
these examples are discussed in \refsec{x-learning}.

\textbf{Compositionality.}
From each example,
we compute the logical form size (number of rules applied)
of the highest-scoring candidate that
executes to the correct denotation.
The histogram in \reffig{predCount}
shows that a significant number of logical forms are non-trivial.

\textbf{Beam size and pruning.}
\reffig{beam} shows the results
with and without pruning on various beam sizes.
Apart from saving time,
pruning also prevents bad logical forms from
clogging up the beam
which hurts both oracle and accuracy metrics.

\subsection{Features}\label{sec:x-features}

\textbf{Effect of features.}
\reftab{x-ablation}(b) shows the accuracy
when some feature types are ablated.
The most influential features are lexicalized phrase-predicate features,
which capture the relationship between phrases and logical operations
(e.g., relating \emph{``last''} to $\T{argmax}$)
as well as between phrases and relations
(e.g., relating \emph{``before''} to $\T{<}$ or $\T{Next}$,
and relating \emph{``who''} to the relation $\T{Name}$).

\textbf{Anchoring with trigger words.}
In our parsing algorithm,
relations and logical operations are not anchored to the utterance.
We consider an alternative approach where
logical operations are anchored to
``trigger'' phrases,
which are hand-coded based on co-occurrence statistics
(e.g., we trigger a $\T{count}$ logical form
with \emph{how}, \emph{many}, and \emph{total}).

\reftab{x-ablation}(c) shows that
the trigger words do not significantly impact the accuracy,
suggesting that the original system is already able to learn the relationship
between phrases and operations
even without a manual lexicon.
As an aside, the huge drop in oracle
is because fewer ``semantically incorrect'' logical forms are generated;
we discuss this phenomenon in the next subsection.
 
\FigTop{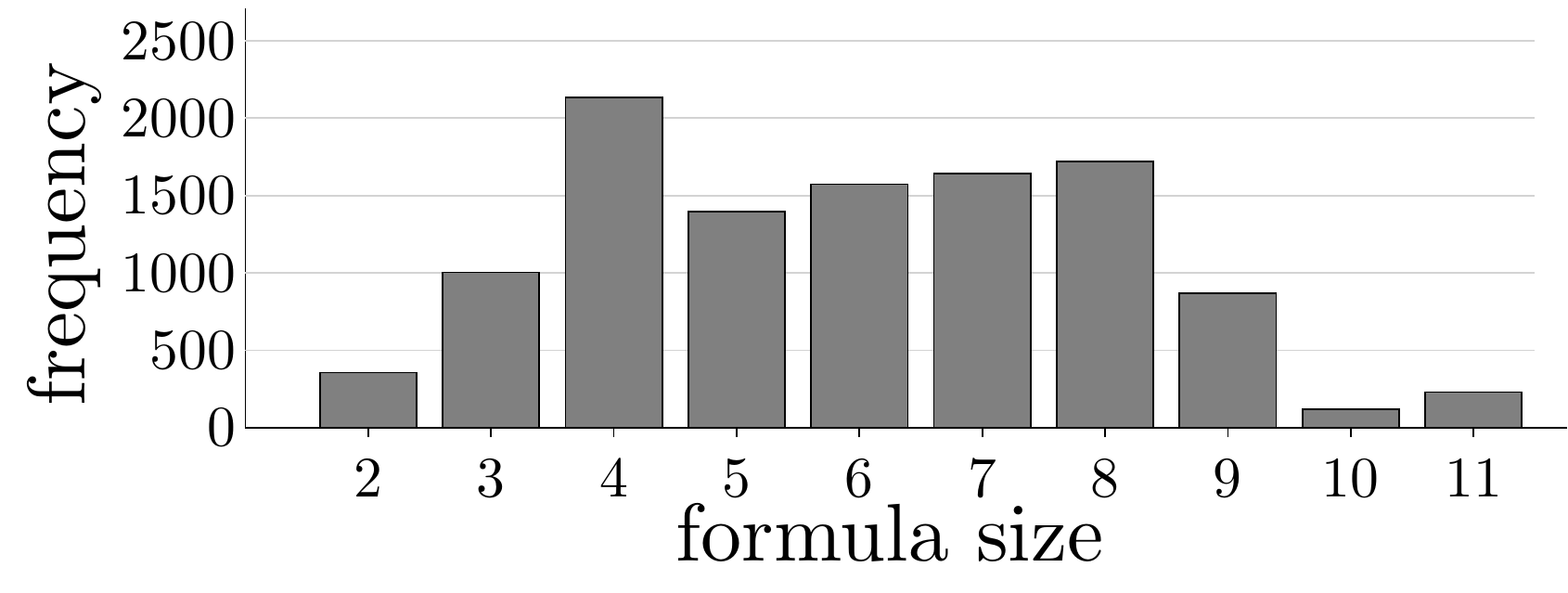}{0.35}{predCount}
{Sizes of the highest-scoring correct candidate logical forms in development examples.}
\FigTop{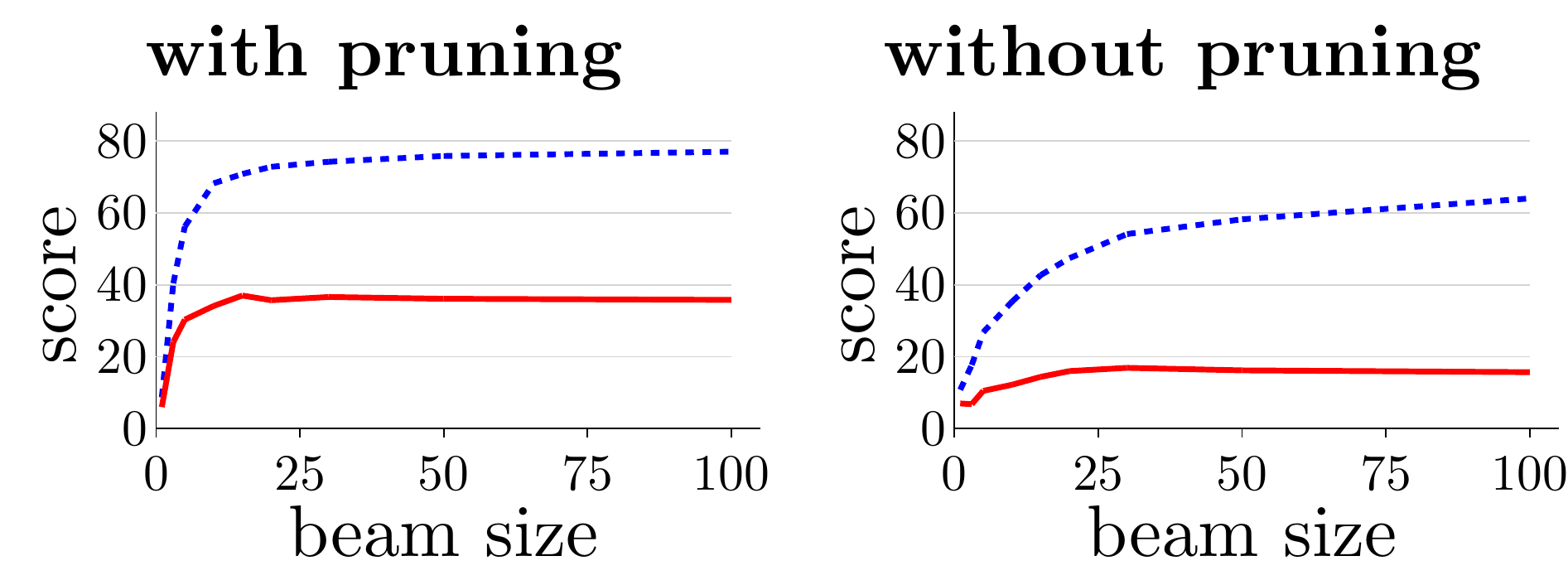}{0.35}{beam}
{Accuracy (solid red) and oracle (dashed blue) scores with different beam sizes.}

\subsection{Semantically correct logical forms}\label{sec:x-learning}
In our setting,
we face a new challenge that arises from learning with denotations:
with deeper compositionality,
a larger number of
nonsensical logical forms
can execute to the correct denotation.
For example, if the target answer is a small number (say, 2),
it is possible to count the number of rows with some random properties
and arrive at the correct answer.
However, as the system encounters more examples,
it can potentially learn to disfavor them
by recognizing the characteristics of semantically correct logical forms.

\textbf{Generating semantically correct logical forms.}
The system can learn the features of semantically correct logical forms
only if it can generate them in the first place.
To see how well the system can generate correct logical forms,
looking at the oracle score is insufficient
since bad logical forms can execute to the correct denotations.
Instead, we randomly chose 200 examples and manually annotated them with logical forms
to see if a trained system can produce the annotated logical form as a candidate.

Out of 200 examples, we find that 79\% can be manually annotated.
The remaining ones include artifacts such as unhandled question types
(e.g., yes-no questions, or questions with phrases \emph{``same''} or \emph{``consecutive''}),
table cells that require advanced normalization methods
(e.g., cells with comma-separated lists),
and incorrect annotations.

The system generates the annotated logical form among the candidates in 53.5\%
of the examples.
The missing examples are mostly caused by anchoring errors
due to lexical mismatch
(e.g., $\emph{``Italian''} \to \T{Italy}$,
or \emph{``no zip code''} $\to$ an empty cell in the zip code column)
or the need to generate complex logical forms from a single phrase
(e.g., $\emph{``May 2010''} \to
\T{>=}.\C{2010-05-01} \sqcap \T{<=}.\C{2010-05-31}$).

\subsection{Error analysis}\label{sec:x-error}
The errors on the development data
can be divided into four groups.
The first two groups are unhandled question types (21\%)
and the failure to anchor entities (25\%) as described in \refsec{x-learning}.
The third group is normalization and type errors (29\%):
although we handle some forms of entity normalization,
we observe many unhandled string formats such as times (e.g., \emph{3:45.79})
and city-country pairs (e.g., \emph{Beijing, China}),
as well as complex calculation such as computing time periods
(e.g., \emph{12pm--1am} $\to$ 1 hour).
Finally, we have ranking errors (25\%) which mostly occur when
the utterance phrase and the relation are obliquely related
(e.g., \emph{``airplane''} and $\T{Model}$).

%% file: discussion.tex
\section{Discussion}
Our work simultaneously increases
the breadth of knowledge source
and the depth of compositionality
in semantic parsing.
This section explores the connections
in both aspects to related work.

\textbf{Logical coverage.}
Different semantic parsing systems
are designed to handle different sets of logical operations
and degrees of compositionality.
For example, form-filling systems \cite{wang2011semantic}
usually cover a smaller scope of operations and compositionality,
while early statistical semantic parsers
for question answering
\cite{wong07synchronous,zettlemoyer07relaxed}
and high-accuracy natural language interfaces for databases
\cite{androutsopoulos95nlidb,popescu03precise} 
target more compositional utterances
with a wide range of logical operations.
This work aims to increase the logical coverage even further.
For example, compared to the Geoquery dataset,
the \dataset dataset includes a move diverse set of logical operations,
and while it does not have extremely compositional questions like in Geoquery
(e.g., \emph{``What states border states that border states that border Florida?''}),
our dataset contains fairly compositional questions on average.

To parse a compositional utterance,
many  works rely on a lexicon
that translates phrases to
entities, relations, and logical operations.
A lexicon can be
automatically generated
\cite{unger2011pythia,unger2012template},
learned from data
\cite{zettlemoyer07relaxed,kwiatkowski11lex},
or extracted from external sources
\cite{cai2013large,berant2013freebase},
but requires some techniques to generalize to unseen data.
Our work takes a different approach
similar to the logical form growing algorithm
in \newcite{berant2014paraphrasing}
by not anchoring relations and operations to the utterance.

\textbf{Knowledge domain.}
Recent works on semantic parsing for question answering
operate on more open and diverse data domains.
In particular, large-scale knowledge bases have gained popularity
in the semantic parsing community
\cite{cai2013large,berant2013freebase,fader2014open}.
The increasing number of relations and entities
motivates new resources and techniques for improving the accuracy,
including the use of
ontology matching models \cite{kwiatkowski2013scaling},
paraphrase models \cite{fader2013paraphrase,berant2014paraphrasing},
and unlabeled sentences \cite{krishnamurthy2013jointly,reddy2014large}.

Our work leverages open-ended data from the Web
through semi-structured tables.
There have been several studies on analyzing or inferring the table schemas
\cite{cafarella2008webtables,venetis2011recovering,syed2010exploiting,limaye2010annotating}
and answering search queries by
joining tables on similar columns
\cite{cafarella2008webtables,gonzalez2010google,pimplikar2012answering}.
While the latter is similar to question answering,
the queries tend to be keyword lists instead of natural language sentences.
In parallel,
open information extraction \cite{wu2010open,masaum2012open}
and knowledge base population \cite{ji2011knowledge}
extract information from web pages
and compile them into structured data.
The resulting knowledge base is systematically organized,
but as a trade-off, 
some knowledge is inevitably lost during extraction
and the information is forced to conform to a specific schema.
To avoid these issues, we choose to work on HTML tables directly.

In future work, we wish to draw information from
other semi-structured formats such as 
colon-delimited pairs \cite{wong2009scalable},
bulleted lists \cite{gupta2009answering}, and
top-$k$ lists \cite{zhang2013automatic}.
\newcite{pasupat2014extraction} used a framework similar to ours to extract
entities from web pages, where the ``logical forms'' were XPath
expressions.  A natural direction is to combine the logical compositionality of this work
with the even broader knowledge source of general web pages.

%% file: acknowledgement.tex
\paragraph{Acknowledgements.}
We gratefully acknowledge the support of
the Google Natural Language Understanding Focused Program
and the Defense Advanced Research Projects Agency (DARPA)
Deep Exploration and Filtering of Text (DEFT) Program
under Air Force Research Laboratory (AFRL)
contract no.\ FA8750-13-2-0040.

\paragraph{Data and reproducibility.}
The \dataset dataset can be downloaded at
{\small \url{http://nlp.stanford.edu/software/sempre/wikitable/}}.
Additionally, code, data, and experiments for this paper are available on the CodaLab platform at
{\small \url{https://www.codalab.org/worksheets/0xf26cd79d4d734287868923ad1067cf4c/}}.